\documentclass[nonatbib]{article}

\usepackage[final]{neurips_2021}

\newcommand{\citet}{\cite}
\newcommand{\citep}{\cite}

\usepackage[utf8]{inputenc} %
\usepackage[T1]{fontenc}    %
\usepackage{hyperref}       %
\usepackage{url}            %
\usepackage{booktabs}       %
\usepackage{amsfonts}       %
\usepackage{nicefrac}       %
\usepackage{microtype}      %
\usepackage{xcolor}         %

\definecolor{Mycolor1}{cmyk}{0.1, 0.7808, 0.4429, 0.1412}
\definecolor{mygreen}{rgb}{0.35, 0.5, 0.0}

\usepackage{amsmath}
\usepackage{braket}

\usepackage{tikz}
\usetikzlibrary{decorations.pathmorphing}
\tikzset{snake it/.style={decorate, decoration=snake,segment length=1.5mm}}

\newenvironment{diagram}
{
\begin{tikzpicture}[baseline = (X.base),every node/.style={scale=0.7},scale=.55]
}
{
\end{tikzpicture}
}

\usepackage{caption}
\usepackage{subcaption}

\title{Distributive Pre-Training of Generative Modeling Using Matrix Product States}

\author{%
  Sheng-Hsuan Lin \thanks{shenghsuan.lin@tum.de}
  \\
  Department of Physics\\
  Technische Universit{\"a}t M{\"u}nchen\\
  85748 Garching, Germany
  \And
  Olivier Kuijpers \\
  Department of Physics\\
  Technische Universit{\"a}t M{\"u}nchen\\
  85748 Garching, Germany
   \AND
   Sebastian Peterhansl  \\
  Department of Physics\\
  Technische Universit{\"a}t M{\"u}nchen\\
  85748 Garching, Germany
   \And
   Frank Pollmann\\
  Department of Physics\\
  Technische Universit{\"a}t M{\"u}nchen\\
  85748 Garching, Germany
}

\begin{document}

\maketitle

\begin{abstract}
Tensor networks have recently found applications in machine learning for both supervised learning and unsupervised learning.
The most common approaches for training these models are gradient descent methods.
In this work, we consider an alternative training scheme utilizing basic tensor network operations, e.g., summation and compression.
The training algorithm is based on compressing the superposition state constructed from all the training data in product state representation.
The algorithm could be parallelized easily and only iterates through the dataset once. Hence, it serves as a pre-training algorithm.
We benchmark the algorithm on the MNIST dataset and show reasonable results for generating new images and classification tasks.
Furthermore, we provide an interpretation of the algorithm as a compressed quantum kernel density estimation for the probability amplitude of input data.
\end{abstract}

\section{\label{sec: Intro} Introduction}

Machine learning and many-body physics have great similarities in the studies of finding low-dimensional and meaningful representations over exponentially large degrees of freedom~\citep{bengio2013representation,goodfellow2016deep}.
Tensor networks have been proven to be efficient representations for quantum many-body systems with low entanglement~\citep{hastings2007area,eisert2010colloquium}.
With numerical algorithms, tensor network states (TNSs) are parameterized models that could be optimized variationally to solve many-body problems~\citep{schollwock2011density,bridgeman2017hand}, e.g. searching the ground state.
As parameterized models, TNSs have attracted interest in recent years for solving supervised learning problems~\citep{stoudenmire2016supervised} and also unsupervised learning problems~\citep{PhysRevX.8.031012}.
Being a promising method, tensor networks may also give us insight into machine learning problems.

In this work, we propose a new pre-training algorithm for models based on matrix-product states (MPSs) for unsupervised generative modeling and supervised learning on the MNIST dataset~\citep{lecun2010mnist}.
The algorithm is an iterative compression over the summation of quantum states~\footnote{The uncompressed superposition states are considered previously by Martyn et al.~\cite{martyn2020entanglement}, where they showed the uncompressed states form good models but concluded these states are not relevant for tensor network models because of the high entanglement.} encoding the input data, which could be parallelized as a tree-based reduction algorithm and run distributively.

The main contributions of our article are:
(i) We propose a simple pre-training algorithm for MPS models that could be parallelized and run distributively with potentially exponential speedup.
(ii) We identify the combination of the orthonormal feature map~\citep{stoudenmire2016supervised} and the superposition state of all the quantum states encoding data as a quantum version of the kernel density estimation.
(iii) We propose a sampling algorithm for continuous variables by viewing the combination of wavefunctions over discrete variables ${\bf s}$ and the orthonormal feature maps as quantum latent variable wavefunctions.

The remainder of the paper is organized as follows:
We introduce the tensor network notation for continuous variables and review the idea of feature maps in Sec.~\ref{sec: feature_map}.
We discuss the MPS-based Born machines for both discrete and continuous variables in Sec.~\ref{sec: models}.
The proposed pre-training algorithm is shown in Sec.~\ref{sec: algorithm}. 
We show the result in Sec.~\ref{sec: Result} and discuss the implication in Sec.~\ref{sec: discussion}.

\section{\label{sec: Method} Method}

\subsection{Tensor network notation and feature map\label{sec: feature_map}}

We introduce some non-conventional tensor network notations for continuous variables to facilitate the discussion.
In the standard tensor network notation, a straight line is associated with a discrete index, for example, $s$. Here, we represent a continuous degree of freedom, for example, $x\in\mathbb{R}$ or $\mathbb{C}$, by a curly line.
Similar to representing a vector in standard tensor network notation, the univariate function $f$ is denoted by a rounded square box with a curly leg
\begin{equation} 
\begin{diagram}
\draw[rounded corners] (1,2) rectangle (2,1);
\draw (1.5,1.5) node (X) {$f$};
\draw[draw=blue, snake it] (1.5,2) -- (1.5,3);
\draw (1.5,3.3) node {$x$};
\end{diagram} =  f(x) .
\end{equation}

By this convention, the Dirac-delta function $\delta(x-\xi)$ for $x, \xi\in\mathbb{R}$ is denoted by the curly line
\begin{equation} 
\begin{diagram}
\draw (1.5,1.7) node {$\xi$};
\draw (1.5,2.5) node (X) {$\phantom{X}$};
\draw (1.5,3.3) node {$x$};
\draw[draw=blue, snake it] (1.5,2) -- (1.5,3); 
\end{diagram} = \delta(x-\xi) ,
\end{equation}
which is similar to the Kronecker delta represented by the straight line.
Connecting the curly lines means taking the integral over the continuous degree of freedom by some measure $\mu(x)$.
Similarly, connecting the straight lines represents a summation of the discrete degrees of freedom.

We apply the same notation for the local feature map introduced by~\citet{stoudenmire2016supervised}.
The local feature map is a vector-valued function defined as
\begin{equation}
    \phi^{s}(x) = \begin{pmatrix} \phi^{s=0}(x) \\ \vdots \\ \phi^{s=d-1}(x) \end{pmatrix} \in \mathbb{C}^d
\qquad \text{denoted as} \qquad 
\begin{diagram}
\draw[rounded corners] (1,2) rectangle (2,1);
\draw (1.5,1.5) node (X) {$\phi$}; 
\draw (1.5,1) -- (1.5,.5); 
\draw (1.5,0.2) node {$s$};
\draw[draw=blue, snake it] (1.5,2) -- (1.5,2.5);
\draw (1.5,2.8) node {$x$};
\end{diagram} = \phi^{s}(x).
\end{equation}

A local feature map is orthonormal if it satisfies the following condition.
\begin{equation} \label{eq: orthogonality_condition}
    \int \bar{\phi}^{s}(x) \phi^{s'}(x) d\mu(x) = \delta_{s s'}
\qquad \text{denoted as} \qquad 
\begin{diagram}
\draw[rounded corners] (1,1.5) rectangle (2,.5);
\draw[rounded corners] (1,-1.5) rectangle (2,-.5);
\draw (1.5,1) node {$\bar{\phi}$}; \draw (1.5,-1) node (X) {$\phi$};
\draw[draw=blue, snake it] (1.5,.5) -- (1.5,-.5); 
\draw (1.5,-1.5) -- (1.5,-2.); \draw (1.5,1.5) -- (1.5,2.);
\draw (1.5,-2.3) node {$s'$};
\draw (1.5,2.3) node {$s$};
\draw (1.5,0) node (X) {$\phantom{X}$};
\end{diagram}\;
= 
\begin{diagram}
\draw (1.5,1.) -- (1.5,-1.);
\draw (1.5,0) node (X) {$\phantom{X}$};
\draw (1.5,-1.3) node {$s'$};
\draw (1.5,1.3) node {$s$};
\end{diagram}
= \delta_{s s'}
\end{equation}

A local feature map acting on a single continuous variable $x$ creates a normalized discrete wavefunction if the local feature map satisfies the condition 
\begin{equation} \label{eq: normalized}
\sum_s \lvert \phi^s(x) \rvert^2  = 1
\end{equation}

Note that this does not imply and is different from the resolution of identity $\sum_{s} \phi^{s}(x)\bar{\phi}^{s}(x')  = \delta(x-x')$. That is, in general,
\begin{equation} 
\begin{diagram}
\draw[rounded corners] (1,1.5) rectangle (2,.5);
\draw[rounded corners] (1,-1.5) rectangle (2,-.5);
\draw (1.5,1) node {${\phi}$}; \draw (1.5,-1) node (X) {$\bar{\phi}$};
\draw (1.5,.5) -- (1.5,-.5); 
\draw[draw=blue, snake it] (1.5,-1.5) -- (1.5,-2.);
\draw[draw=blue, snake it] (1.5,1.5) -- (1.5,2.);
\draw (1.5,0) node (X) {$\phantom{X}$};
\draw (1.5,2.3) node {$x$};
\draw (1.5,-2.3) node {$x'$};
\end{diagram}\;
\neq
\begin{diagram}
\draw[draw=blue, snake it] (1.5,1.) -- (1.5,-1.);
\draw (1.5,0) node (X) {$\phantom{X}$};
\draw (1.5,1.3) node {$x$};
\draw (1.5,-1.3) node {$x'$};
\end{diagram}
= \delta(x-x').
\end{equation}
This is because the resolution of identity only holds when the local feature map is a complete orthonormal basis set of functions.
Such a feature map would be infinite-dimensional and is not considered in practice.

Here we give some examples of the known local feature maps:\\
\emph{Example 1:} $\phi(x) = [\cos(\frac{\pi}{2}x), \sin(\frac{\pi}{2}x)],\  x\in[0,1]$
satisfies the condition in Eq.~\eqref{eq: normalized} but not the orthonormal condition in Eq.~\eqref{eq: orthogonality_condition}.

\emph{Example 2:}  $\phi(x) = [e^{i(3\pi/2)x}\cos(\frac{\pi}{2}x), e^{-i(3\pi/2)x}\sin(\frac{\pi}{2}x)], x\in[0,1]$
satisfies the condition in Eq.~\eqref{eq: normalized} and the orthonormal condition in Eq.~\eqref{eq: orthogonality_condition}.

\emph{Example 3:}  $\phi(x) = [\text{sgn}(x)-\text{sgn}(x-0.5), \text{sgn}(x-0.5)-\text{sgn}(x-1)], x\in[0,1]$
satisfies the condition in Eq.~\eqref{eq: normalized} and the orthonormal condition in Eq.~\eqref{eq: orthogonality_condition}.

The (global) feature map over variables is often taken to be the tensor product of local feature maps. 
\begin{equation}
    \Phi^{\bf s}({\bf x}) = \Phi^{s_1, s_2, \ldots s_N}({\bf x}) = \phi^{s_1}(x_1)\otimes \phi^{s_2}(x_2) \otimes \cdots \otimes \phi^{s_N}(x_N) .
\end{equation}
The feature map maps any single input data of continuous variables ${\bf x}^{(i)}$ to a normalized wavefunction $\Phi^{\bf s}({\bf x}^{(i)})$ over the discrete variables ${\bf s}$ if the local feature map satisfies the condition in Eq.~\eqref{eq: normalized}.
We can re-express the description as
\begin{align}
    \bar{\Phi}^{\bf s}({\bf x}^{(i)}) &= \int \bar{\Phi}^{\bf s}({\bf x}) \delta({\bf x}-{\bf x}^{(i)}) d{\bf x} \\
    &= 
\begin{diagram}
\draw (1.5,-1.5) node {$x^{(i)}_1$};
\draw (3.5,-1.5) node {$x^{(i)}_2$};
\draw (5.5,-1.5) node {$x^{(i)}_3$};
\draw (7.5,-1.5) node {$x^{(i)}_4$};
\draw (9.5,-1.5) node {$x^{(i)}_5$};
\draw[rounded corners] (1,0.5) rectangle (2,-0.5);
\draw (1.5,0) node {$\bar{\phi}$};
\draw[rounded corners] (3,0.5) rectangle (4,-0.5);
\draw (3.5,0) node {$\bar{\phi}$};
\draw[rounded corners] (5,0.5) rectangle (6,-0.5);
\draw (5.5,0) node {$\bar{\phi}$};
\draw (1.5,1) -- (1.5,0.5);
\draw (3.5,1) -- (3.5,0.5);
\draw (5.5,1) -- (5.5,0.5);
\draw (7.5,1) -- (7.5,0.5);
\draw (9.5,1) -- (9.5,0.5);
\draw (1.5,1.3) node {$s_1$};
\draw (3.5,1.3) node {$s_2$};
\draw (5.5,1.3) node {$s_3$};
\draw (7.5,1.3) node {$s_4$};
\draw (9.5,1.3) node {$s_5$};
\draw[rounded corners] (7,0.5) rectangle (8,-0.5);
\draw (7.5,0) node {$\bar{\phi}$};
\draw[rounded corners] (9,0.5) rectangle (10,-0.5);
\draw (9.5,0) node {$\bar{\phi}$};
\draw[draw=blue, snake it] (1.5,-1) -- (1.5,-0.5);
\draw[draw=blue, snake it] (3.5,-1) -- (3.5,-0.5);
\draw[draw=blue, snake it] (5.5,-1) -- (5.5,-0.5);
\draw[draw=blue, snake it] (7.5,-1) -- (7.5,-0.5);
\draw[draw=blue, snake it] (9.5,-1) -- (9.5,-0.5);
\end{diagram} \nonumber 
\ \ =\ \ 
\begin{diagram}
\draw[rounded corners] (0.8,0.5) rectangle (2.2,-0.5);
\draw (1.5,0) node {$\bar{\phi}(x^{(i)}_1)$};
\draw[rounded corners] (2.8,0.5) rectangle (4.2,-0.5);
\draw (3.5,0) node {$\bar{\phi}(x^{(i)}_2)$};
\draw[rounded corners] (4.8,0.5) rectangle (6.2,-0.5);
\draw (5.5,0) node {$\bar{\phi}(x^{(i)}_3)$};
\draw (1.5,1) -- (1.5,0.5);
\draw (3.5,1) -- (3.5,0.5);
\draw (5.5,1) -- (5.5,0.5);
\draw (7.5,1) -- (7.5,0.5);
\draw (9.5,1) -- (9.5,0.5);
\draw (1.5,1.3) node {$s_1$};
\draw (3.5,1.3) node {$s_2$};
\draw (5.5,1.3) node {$s_3$};
\draw (7.5,1.3) node {$s_4$};
\draw (9.5,1.3) node {$s_5$};
\draw[rounded corners] (6.8,0.5) rectangle (8.2,-0.5);
\draw (7.5,0) node {$\bar{\phi}(x^{(i)}_4)$};
\draw[rounded corners] (8.8,0.5) rectangle (10.2,-0.5);
\draw (9.5,0) node {$\bar{\phi}(x^{(i)}_5)$}; 
\end{diagram} \nonumber.
\end{align}
That is we can think of the normalized wavefunction $\Phi^{\bf s}({\bf x}^{(i)})$ on discrete variables ${\bf s}$ as the feature map $\bar{\Phi}^{\bf s}({\bf x})$ acting on the input $\delta({\bf x}-{\bf x}^{(i)})$.
With the same formalism, one could also map any wavefunction $\Psi({\bf x})$ over continuous variables ${\bf x}$ to a wavefunction $\Psi^{\bf s}$ over discrete variables ${\bf s}$ by
\begin{align}
    &\int \bar{\Phi}^{\bf s}({\bf x}) \Psi({\bf x}) d\mu({\bf x}) = \Psi^{\bf s} 
    =
\begin{diagram}
\draw[rounded corners] (1,-1) rectangle (10,-2);
\draw[rounded corners] (1,0.5) rectangle (2,-0.5);
\draw (1.5,0) node {$\bar{\phi}$};
\draw[rounded corners] (3,0.5) rectangle (4,-0.5);
\draw (3.5,0) node {$\bar{\phi}$};
\draw[rounded corners] (5,0.5) rectangle (6,-0.5);
\draw (5.5,0) node {$\bar{\phi}$};
\draw (5.5,-1.5) node {$\Psi$};
\draw (1.5,1) -- (1.5,0.5);
\draw (3.5,1) -- (3.5,0.5);
\draw (5.5,1) -- (5.5,0.5);
\draw (7.5,1) -- (7.5,0.5);
\draw (9.5,1) -- (9.5,0.5);
\draw (1.5,1.3) node {$s_1$};
\draw (3.5,1.3) node {$s_2$};
\draw (5.5,1.3) node {$s_3$};
\draw (7.5,1.3) node {$s_4$};
\draw (9.5,1.3) node {$s_5$};
\draw[rounded corners] (7,0.5) rectangle (8,-0.5);
\draw (7.5,0) node {$\bar{\phi}$};
\draw[rounded corners] (9,0.5) rectangle (10,-0.5);
\draw (9.5,0) node {$\bar{\phi}$};
\draw[draw=blue, snake it] (1.5,-1) -- (1.5,-0.5);
\draw[draw=blue, snake it] (3.5,-1) -- (3.5,-0.5);
\draw[draw=blue, snake it] (5.5,-1) -- (5.5,-0.5);
\draw[draw=blue, snake it] (7.5,-1) -- (7.5,-0.5);
\draw[draw=blue, snake it] (9.5,-1) -- (9.5,-0.5);
\end{diagram} .
\end{align}
Note, however, the resulting wavefunction $\Psi^{\bf s}$ is not normalized since the local feature map is in general not the resolution of identity.

The feature map could map any normalized wavefunction $\Psi^{\bf s}$ to a normalized wavefunction $\Psi({\bf x})$ by
\begin{align} \label{eq: discrete_wf_w_feature_map}
\sum_{\bf s} \Phi^{\bf s}({\bf x})  \Psi^{\bf s}  = \Psi({\bf x})
=
\begin{diagram}
\draw[rounded corners] (1,-1) rectangle (10,-2);
\draw[rounded corners] (1,0.5) rectangle (2,-0.5);
\draw (1.5,0) node {${\phi}$};
\draw[rounded corners] (3,0.5) rectangle (4,-0.5);
\draw (3.5,0) node {${\phi}$};
\draw[rounded corners] (5,0.5) rectangle (6,-0.5);
\draw (5.5,0) node {${\phi}$};
\draw (5.5,-1.5) node {$\Psi$};
\draw[draw=blue, snake it] (1.5,1) -- (1.5,0.5);
\draw[draw=blue, snake it] (3.5,1) -- (3.5,0.5);
\draw[draw=blue, snake it] (5.5,1) -- (5.5,0.5);
\draw[draw=blue, snake it] (7.5,1) -- (7.5,0.5);
\draw[draw=blue, snake it] (9.5,1) -- (9.5,0.5);
\draw (1.5,1.3) node {$x_1$};
\draw (3.5,1.3) node {$x_2$};
\draw (5.5,1.3) node {$x_3$};
\draw (7.5,1.3) node {$x_4$};
\draw (9.5,1.3) node {$x_5$};
\draw[rounded corners] (7,0.5) rectangle (8,-0.5);
\draw (7.5,0) node {${\phi}$};
\draw[rounded corners] (9,0.5) rectangle (10,-0.5);
\draw (9.5,0) node {${\phi}$};
\draw (1.5,-1) -- (1.5,-0.5);
\draw (3.5,-1) -- (3.5,-0.5);
\draw (5.5,-1) -- (5.5,-0.5);
\draw (7.5,-1) -- (7.5,-0.5);
\draw (9.5,-1) -- (9.5,-0.5);
\end{diagram} ,
\end{align}
if the orthonormal condition Eq.~\eqref{eq: orthogonality_condition} holds.
Examples of such orthonormal mappings are given above.

An important and elementary expression that will appear repeatedly is  
\begin{align} 
\begin{diagram}
\draw[rounded corners] (1,1.5) rectangle (2,.5);
\draw[rounded corners] (1,-1.5) rectangle (2,-.5);
\draw (1.5,1) node {$\phi$}; \draw (1.5,-1) node (X) {$\bar{\phi}$};
\draw (1.5,.5) -- (1.5,-.5); 
\draw[draw=blue, snake it] (1.5,-1.5) -- (1.5,-2.);
\draw[draw=blue, snake it] (1.5,1.5) -- (1.5,2.);
\draw (1.5,-2.3) node {$\xi$};
\draw (1.5,2.3) node {$x$};
\draw (1.5,0) node (X) {$\phantom{X}$};
\end{diagram}\;
= \sum_s \phi^{s}({x}) \bar{\phi^{s}}(\xi)  = \sum_s \int \phi^{s}({x}) \bar{\phi^{s}}({x'})\delta({x'}-\xi) d{x'}  ,
\end{align}
which can be interpreted as the projection of the Dirac-delta function at $\xi$ on a finite basis set $\phi^s(x)$.
Therefore, this expression represents an approximation or smoothing of Dirac-delta functions peaked at $x=\xi$
(see Fig.~\ref{fig: basis_set} in Appendix~\ref{appendix: data}) .
We will see from this perspective that the summation of quantum states encoding the data points is related to the summation of smoothed Dirac-delta functions over the given data points.

\subsection{Generative Models: Born Machine based on MPS and feature maps \label{sec: models}}

\paragraph{Born machines:}
Quantum wavefunctions could be utilized as machine learning tools describing the probability distribution given by Born's rule, i.e., $P=|\psi|^2$.
The Born machines~\citep{cheng2018information} are parameterized quantum wavefunctions, e.g. tensor network states~\citep{PhysRevX.8.031012,cheng2019tree} or parameterized quantum circuits~\citep{liu2018differentiable}, that are applied as generative models over discrete variables for learning  the probability distributions of images.
Born machines based on MPSs have the advantage that they give a tractable likelihood, i.e., the probability amplitude can be evaluated efficiently and allows ancestral sampling~\citep{ferris2012perfect}.

\paragraph{Born machine based on MPS over continuous variables:}
In this work, we consider Born machines based on MPSs for both discrete and continuous variables.
MPSs could also parameterize wavefunctions over continuous variables with the help of orthonormal feature maps~\citep{stoudenmire2016supervised}.
We see this from Eq.~\eqref{eq: discrete_wf_w_feature_map} by replacing the $\Psi^{\bf s}$ by an MPS.
Algorithms for training generative models of discrete variables also work for models with continuous variables~\citep{PhysRevX.8.031012}.
Moreover, the combination of an MPS with a feature map not only defines a model over continuous variables ${\bf x}$, but it also gives an additional useful interpretation as a ``quantum" latent variable model, where the latent variables are the discrete variables ${\bf s}$.
Such an interpretation leads to a sampling algorithm for continuous variables.

\paragraph{Latent variable models:}
 Latent variable models are defined with some latent variables ${\bf z}$ and the parametrized distributions $P_\theta({\bf x}| {\bf z})$ and $P_\theta( {\bf z})$ such that the probability distribution $P({\bf x})$ over the variables ${\bf x}$ can be expressed as $P({\bf x}) = \sum_{\bf z} P_\theta({\bf x}| {\bf z}) P_\theta( {\bf z})$.
Latent variable models can be efficiently sampled by first sampling the latent variables ${\bf z}\sim  P_\theta( {\bf z})$ and then sampling the variables ${\bf x} \sim P_\theta({\bf x}| {\bf z})$ given the latent variables ${\bf z}$.

\paragraph{Latent variable wavefunctions:}
Similarly, we can interpret the MPS with local feature maps as a latent variable wavefunction or a quantum latent variable model. The probability amplitude over the continuous variables ${\bf x}$ is given by the parameterized amplitudes $\Psi_\theta({\bf x}| {\bf s})$ and $\Psi_\theta( {\bf s})$, where ${\bf s}$ are the discrete latent variables, e.g., the spins.
The conditional probability amplitudes need to satisfy the normalization condition, i.e., $ \int \lvert \Psi_\theta({\bf x}| {\bf s})\rvert^2 d{\bf x} = 1, \ \forall\ {\bf s}$, which is automatically satisfied by the feature map $\Phi^{\bf s}({\bf x})$ if the local feature map is orthonormal.
In our setup, we consider a fixed conditional probability amplitude $\Psi({\bf x}| {\bf s}) = \Phi^{\bf s}({\bf x})$.
As a result, we have $\Psi({\bf x}) = \sum_{\bf s} \Phi({\bf x}| {\bf s}) \Psi_\theta( {\bf s})$.

Because the probability of a latent variable wavefunction defines a latent variable model, we can similarly sample the continuous variables ${\bf x}$ according to the probability distribution $| \Psi({\bf x}) |^2$: one can first sample the discrete variables ${\bf s}$ according to ${\bf s} \sim | \Psi_\theta({\bf s}) |^2$ and then sample the continuous variables ${\bf x}$ according to ${\bf x} \sim | \Psi_\theta({\bf x} | {\bf s}) |^2$.

\subsection{Algorithm \label{sec: algorithm}}

Here, we propose a pre-training algorithm for MPS-based Born machines which provides good initialization for bond dimension $\chi$ MPSs that could be applied to generative modeling or supervised learning.
We review the general setup of generative modeling in Appendix~\ref{appendix: generative_modeling}.
The motivation of the algorithm comes from the observation that a special wavefunction, dubbed the digit wavefunction or the sum state $\ket{\Sigma_l}$~\citep{martyn2020entanglement}, captures the data distribution and performs well in the classification task.

The algorithm is based on MPS compression algorithms over this wavefunction.
The digit wavefunction $\ket{\Sigma_l}$ of digit $l$ is defined as the sum over all the training data of digit $l$ in product state representation:
\begin{equation}
    \ket{\Sigma_l} = \Big( \sum_{i; y^{(i)}=l} \ket{\Phi^{\bf s}({\bf x}^{(i)})} \Big) / C_\text{Norm} .
\end{equation}
The $C_\text{Norm}$ is a normalization constant.
It is tempting to consider this state directly as the quantum state defining the probability of the corresponding digit.
As pointed out by~\citet{martyn2020entanglement}, these states have high entanglement and can not be approximated accurately using MPSs with low bond dimensions.
Nevertheless, we consider the pre-training algorithm for MPS-based Born machine with bond dimension $\chi$ as finding the approximated compression of these states.
We denote the uncompressed wavefunction as $\ket{\Sigma_l}$ and the compressed state with bond dimension $\chi$ MPS as $\ket{\Sigma_l^{\chi}}$.
We find that although the compression to low bond dimension MPSs have only small overlaps with the original states, such compressed states are still useful and give reasonable results in both generative modeling and supervised learning tasks.

The state $\ket{\Sigma_l}$ can be constructed by summing up the product states utilizing MPS arithmetic~\citep{schollwock2011density}, which leads to a block-diagonal sparse MPS.
Naively one could first sum up all the product states, then variationally find the optimal truncated MPS of bond dimension $\chi$. We call this method the direct compression.
\begin{equation}
    \ket{\Sigma_l} = \frac{\sum_{i} \Phi^{\bf s}({\bf x}^{(i)})}{C_\text{Norm}}
    \xrightarrow[\text{compression}]{\text{MPS}} \ket{\Sigma_l^{\chi}}
\end{equation}
A more efficient, though approximated, method is the parallel compression.
It is a parallel algorithm with the summation and compression together as a reduction operation.
We first divide the data into $N_\text{data}/\chi$ batches, where each batch can be represented as an MPS of bond dimension $\chi$ exactly.
Then we perform MPS summation and compression between batch $1$ and batch $2$, batch $3$ and batch $4$, etc, and obtain again in the end $N_\text{data}/(2*\chi)$ batches of MPS with bond dimension $\chi$.
Repeating this process, we can sum up the states in tree-like (fan-in) fashion, which has $\log(N_\text{data})$ complexity in time provided enough computation resource.

There are several advantages of using MPSs~\citep{schollwock2011density}:
(i) The addition of two MPSs with bond dimensions $\chi_1, \chi_2$ can be expressed exactly as an MPS with bond dimension $\chi= \chi_1 + \chi_2 $.
(ii) Given an MPS with a higher bond dimension, there are efficient algorithms to find the optimal MPS with a lower bond dimension to approximate the given MPS.
Combining these two properties, the proposed learning algorithm could, in principle, be parallelized.
This is similar to any standard parallel reduction algorithm where MPSs addition and compression are the reduction operation.
This is drastically different to the training using stochastic gradient descent methods which are intrinsically serial and are hard to parallelize.
The algorithm ends when it goes through the dataset exactly once and is suitable for distributive pre-training of generative models.

In the following, we provide an interpretation of the learning algorithm.
Despite the simplicity of the learning process, we show that such an algorithm indeed works in Sec.~\ref{sec: Result}.

\paragraph{Interpretation:}
We first recall the kernel density estimation (KDE) method, which is a class of non-parametric approaches for density estimation.
Given the empirical distribution $\hat{P}({\bf x}) = ({\sum_{i} \delta({\bf x}- {\bf x}^{(i)}) })/{N_\text{data}}$ from the data points, the model is constructed directly from the data points convolving with the kernel, which gives a smoother probability distribution for the data.
\begin{equation} \label{eq: KDE}
    P_\text{KDE}({\bf x}') = \int K({\bf x}'-{\bf x})\hat{P}({\bf x}) d{\bf x} = \int K({\bf x}'-{\bf x})  \frac{{\sum_{i} \delta({\bf x}- {\bf x}^{(i)}) }}{N_\text{data}} d{\bf x} = \sum_i\frac{K({\bf x}'-{\bf x}^{(i)})}{N_\text{data}} \ ,
\end{equation}
where $K({\bf x}'-{\bf x})$ is a kernel function that is non-negative, for example, a Gaussian function.

Given the state $\ket{\Sigma_l}$, one can perform generative modeling or classification based on the evaluation of the overlap $\braket{\Phi^{\bf s}({\bf x})|\Sigma_l}$.
As pointed out in Sec.~\ref{sec: feature_map}, the projection with local feature maps made by a finite basis set will broaden the Dirac-delta function representing the training data.
Therefore, we can interpret the overlap $\braket{\Phi^{\bf s}({\bf x})|\Sigma_l}$ in a different way.
The combination of local feature maps and the summation of the training data is equivalent to performing a summation over smoothed data points to estimate the \emph{probability amplitude} over variables ${\bf x}$. 
To be more precise, we apply the feature map $\Phi^{\bf s}({\bf x})$ on the state $\ket{\Sigma_l}$ and obtain the probability amplitude $\Psi({\bf x}) = \braket{\Phi^{\bf s}({\bf x})|\Sigma_l} = \sum_{(i), s} \braket{{\bf x}|\Phi^{{\bf s}}}\braket{\Phi^{{\bf s}}|{\bf x}^{(i)}} / C_\text{Norm}$ over continuous variables ${\bf x}$.
Notice the similarity to Eq.~\eqref{eq: KDE}.
We reinterpret the overlap $\braket{\Phi^{\bf s}({\bf x})|\Sigma_l}$ as a quantum KDE model estimating the probability amplitude of the input data.
While the quantum KDE requires storing information for all the data points in $\ket{\Sigma_l}$, the compression step in the algorithm acts as finding a compact representation for the models by minimizing the $\mathcal{L}_2$ distance, which resembles the standard approach of minimizing KL-divergence between the model distribution and the empirical distribution.

\section{\label{sec: Result} Results}

\begin{figure}[t]
\centering
\includegraphics[width=0.48\columnwidth]{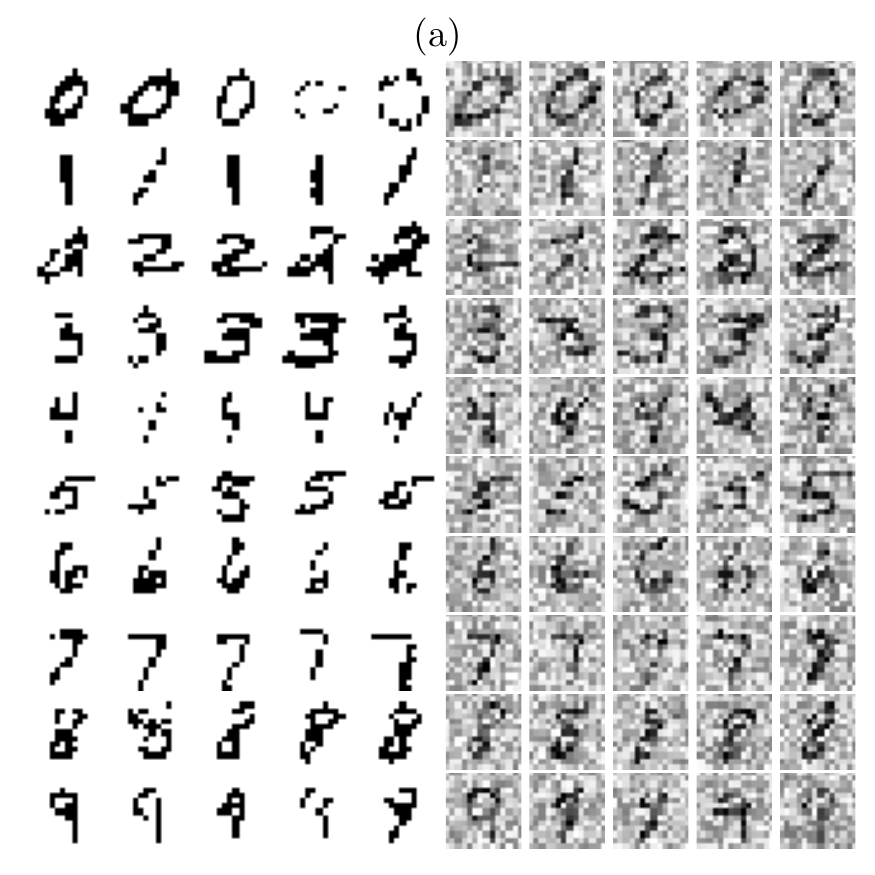}
\includegraphics[width=0.48\columnwidth]{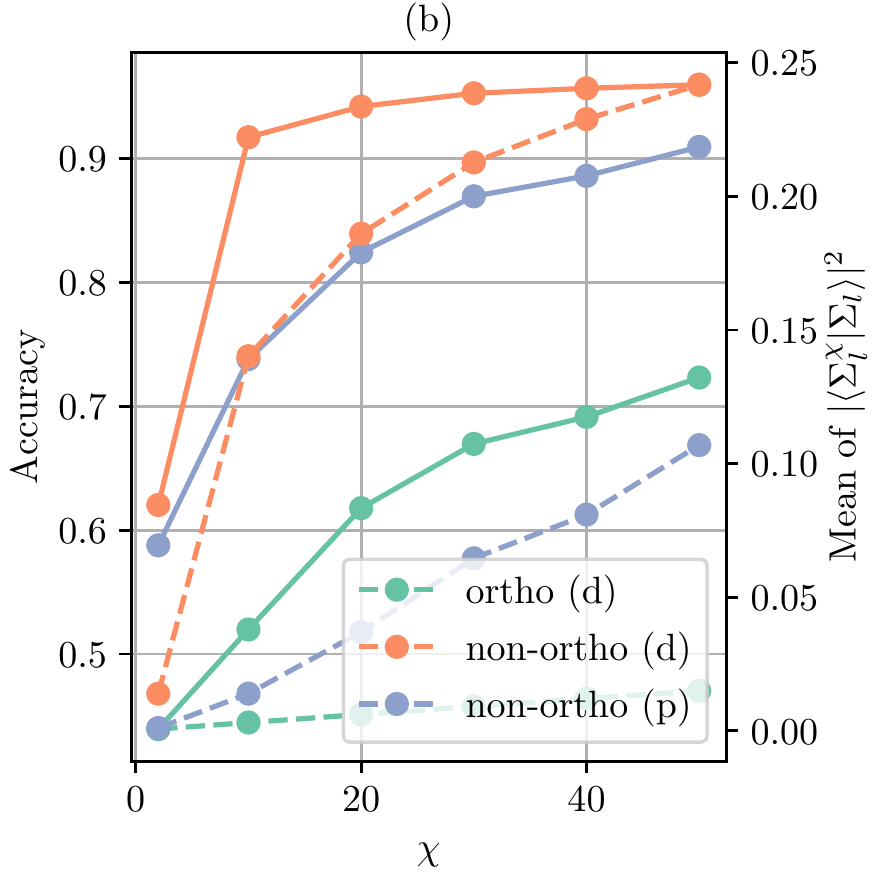}
\caption{
(a) Images sampled from pre-trained MPSs of bond dimension $\chi=50$ by iterative compression scheme.
We sample $5$ binary images and $5$ grey-scaled images from each MPS $\ket{\Sigma^\chi_l}$. 
(b) Test accuracy for MNIST classification by compressed MPS $\ket{\Sigma^\chi_l}$.
We consider the direct compression with orthonormal (green) and the non-orthonormal feature map (orange) and the parallel compression with non-orthonormal feature map (blue).
The solid lines indicate the test accuracy and the dashed lines indicate the mean of the square of overlaps to the exact state $\ket{\Sigma_l}$.
\label{fig: Fig1}
}
\end{figure}

To demonstrate the proposed algorithm, we test the direct compression and iterative compression for obtaining the compressed MPS $\ket{\Sigma^\chi_l}$ for the MNIST dataset. We illustrate the result in Fig.~\ref{fig: Fig1}.

To show that the MPS learning procedure could be a good initialization for generative modeling, we perform an iterative compression of product states encoded by the non-orthonormal local feature map $[\cos(\frac{\pi x}{2}), \sin(\frac{\pi x}{2})]$ to obtain the MPS $\ket{\Sigma^\chi_l}$.
Then we perform ancestral sampling from the MPS wavefunction to get binary images. In addition, we also perform two-stage sampling by first sampling the MPS to get ${\bf s}$ and then sampling the orthonormal local feature map $\phi(x) = [e^{i(3\pi/2)x}\cos(\frac{\pi}{2}x), e^{-i(3\pi/2)x}\sin(\frac{\pi}{2}x)]$ to get the grey-scale images ${\bf x}$. %
The result is shown in Fig.~\ref{fig: Fig1}a.

We use the learned digit wavefunctions to predict the label on test data.
The digit wavefunction, i.e. MPS, with the largest likelihood gives the prediction.
We plot the test accuracy in Fig.~\ref{fig: Fig1}b for direct compression and parallel compression over orthonormal and non-orthonormal local feature maps.

\section{Discussion \label{sec: discussion}}

We propose a pre-training algorithm for MPS models by iterative compression of digit wavefunctions (sum states) $\ket{\Sigma_l}$, which could be parallelized and have potentially exponential speedup.
We test the algorithm on the MNIST dataset and observe reasonable results for tasks including sampling and classification.
We provide a new interpretation of the overlap $\braket{\Phi^{\bf s}({\bf x}) | \Psi_\theta^{\bf s}}$ as the probability amplitude $\Psi_\theta({\bf x})$.
The overlap between the new data point $\ket{\Phi^{\bf s}({\bf x})}$ and the state $\ket{\Sigma_l}$ can then be interpreted as a quantum version of a kernel density estimate for the probability amplitude.

In a recent work~\citet{martyn2020entanglement}, the authors studied the entanglement properties of the uncompressed digit wavefunctions $\ket{\Sigma_l}$ and found a flat Schmidt spectrum.
The authors conclude that the MPS resulting from supervised learning~\citep{stoudenmire2016supervised} must be some different and less entangled states.
Our work is consistent with their observation as we also observe small overlaps with the exact state $\ket{\Sigma_l}$ when using a small bond dimension MPS for compression.
However, surprisingly, these states already lead to good results in supervised learning, e.g., $>95\%$ test accuracy with $\chi=50$ by direct compression.
Moreover, these states capture the correct long-range correlation in the images, as seen from the sampling.
This suggests that the compressed digit wavefunctions could serve as good initial states for both supervised and unsupervised learning, although they have a very small overlap with the exact digit wavefunctions.

It may require further studies to understand whether the method would work for larger datasets.
It would be related to the entanglement and the mutual information scaling of the dataset~\citep{lu2021tensor}, and we envision it would be necessary to use projected entangled pair states (PEPS)~\cite{verstraete2004renormalization,niggemann1997quantum,nishino1998density,sierra1998density} instead of MPS to scale up to larger images.
While in this work, we encode each digit wavefunction $\ket{\Sigma_l}$ as a different MPS, one could also encode all the data in one wavefunction by increasing one index for the labels.
A potential future direction is to incorporate parameterized basis functions for the local feature maps as in~\citep{salimans2017pixelcnn++,hashemi2017chebfun,gorodetsky2019continuous}.

\bibliographystyle{ieeetr}
\bibliography{apssamp}

\appendix

\section{Review for generative modeling \label{appendix: generative_modeling}}

The problem setup for generative modeling is as follows:
We are given a dataset $\{{\bf x}^{(i)}, y^{(i)}\}$, which is a collection of independent and identical distributed samples from the unknown distribution $P({\bf x}, y)$ and $i$ is the index of the samples.
For example, the MNIST dataset consists of images $\{ {\bf x}^{(i)} \}$ and labels $\{ y^{(i)} \}$. 
We represent each image as a vector $\mathbf{x}=(x_1, x_2, \ldots, x_{n_\text{pixel}})$, which is either gray-scaled ${\bf x} \in \mathbb{R}^{n_\text{pixel}}$ or binary $\{0, 1\}^{n_\text{pixel}}$.
Every image has a label $y \in \mathbb{Z}^+$.
The goal of the learning is to obtain an approximation to the true unknown distribution $P({\bf x}, y)$ with some parameterized model $P_\theta({\bf x}, y)$ from the observed data $\{{\bf x}^{(i)}, y^{(i)}\}$.

The generative models can be roughly separated into two categories by whether the models have tractable likelihood, i.e., efficient evaluation of normalized $P({\bf x})$.
The MPS models have tractable likelihood and permits an efficient direct sampling algorithm~\citep{ferris2012perfect}.

When the parameterized model has a tractable likelihood, one common approach for optimizing generative models is to minimize the forward KL-divergence between the empirical distribution $\hat{P}({\bf x}) = \frac{\sum_{i} \delta({\bf x}- {\bf x}^{(i)}) }{C_\text{Norm}}$, where $C_\text{Norm}$ is a normalization constant, and the parameterized model $P_\theta({\bf x})$.
\begin{align}
    \theta_\text{opt} &= \text{argmin}_\theta \int -\log {P_\theta({\bf x})} d\hat{P}({\bf x}) \\
    &= \text{argmin}_\theta  \int -\log {P_\theta({\bf x} )} \sum_{i} \delta({\bf x}- {\bf x}^{(i)}) d\mu(\bf x) \\
&= \text{argmin}_\theta -\sum_{i} \log {P_\theta({\bf x^{(i)}})}
\end{align}

\section{Additional Data \label{appendix: data}}

\begin{figure}[h]
\centering
\includegraphics[width=1.\columnwidth]{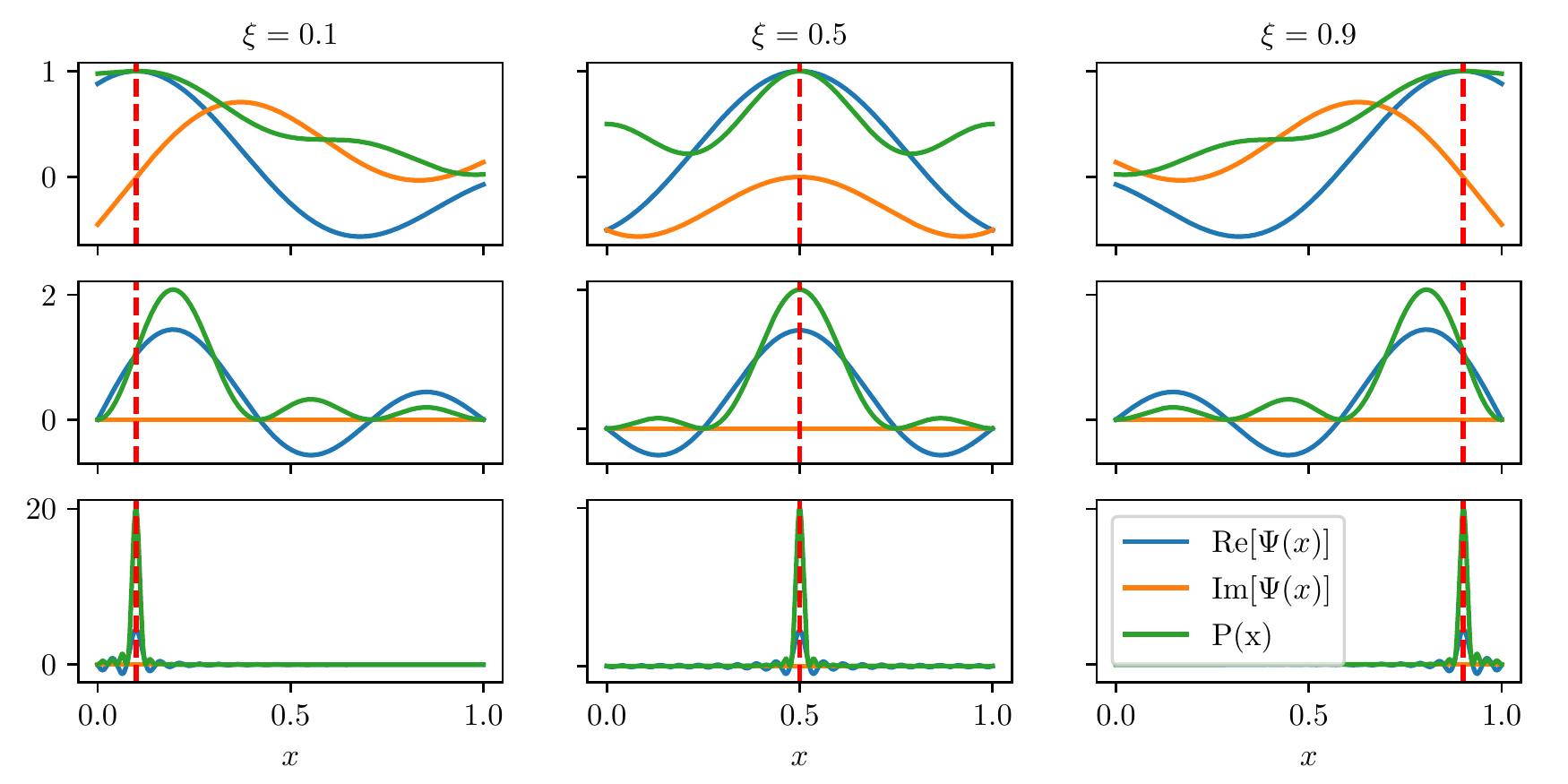}
\caption{Data in the first row shows the $\Psi(x)$ resulting from the orthonormal local feature map $\phi(x) = [e^{i(3\pi/2)x}\cos(\frac{\pi}{2}x), e^{-i(3\pi/2)x}\sin(\frac{\pi}{2}x)]$.
Data in the second row shows the  $\Psi(x)$ resulting from the orthonormal local feature map   $\phi(x) = [\sin(\pi x), \sin(2 \pi x), \sin(3 \pi x), \sin(4\pi x)]$.
Data in the third row shows the  $\Psi(x)$ resulting from the orthonormal local feature map $\phi(x) = [\sin(n\pi x) \ \ n\in\{1,\ldots,40\}]$.
The Dirac-delta functions are denoted by the red dashed lines.
}
\label{fig: basis_set}
\end{figure}

Given a finite orthonormal basis set, the expansion 
\begin{equation*}
\Psi(x) = \sum_s \int \phi^{s}({x}) \bar{\phi^{s}}({x'})\delta({x'}-\xi) d{x'} 
= \sum_s \braket{x|\phi^s} \braket{\phi^s | \xi}    
\end{equation*}
of the Dirac-delta function at $\xi$ would create a broadening of the function.
We show the example in Fig.~\ref{fig: basis_set} considering the different orthonormal local feature maps: (i) $\phi(x) = [e^{i(3\pi/2)x}\cos(\pi x/2), e^{-i(3\pi/2)x}\sin(\pi x/2)]$, (ii)  $\phi(x) = [\sin(\pi x), \sin(2 \pi x), \sin(3 \pi x), \sin(4\pi x)]$, (iii) $\phi(x) = [\sin(n\pi x), \ n\in\{1,\ldots,40\}]$.
We observe that the smoothing behaviour depends on the basis sets and gets closer to the Dirac-delta function with increasing number of basis states.

\begin{figure}[h]
\centering
\includegraphics[width=0.55\columnwidth]{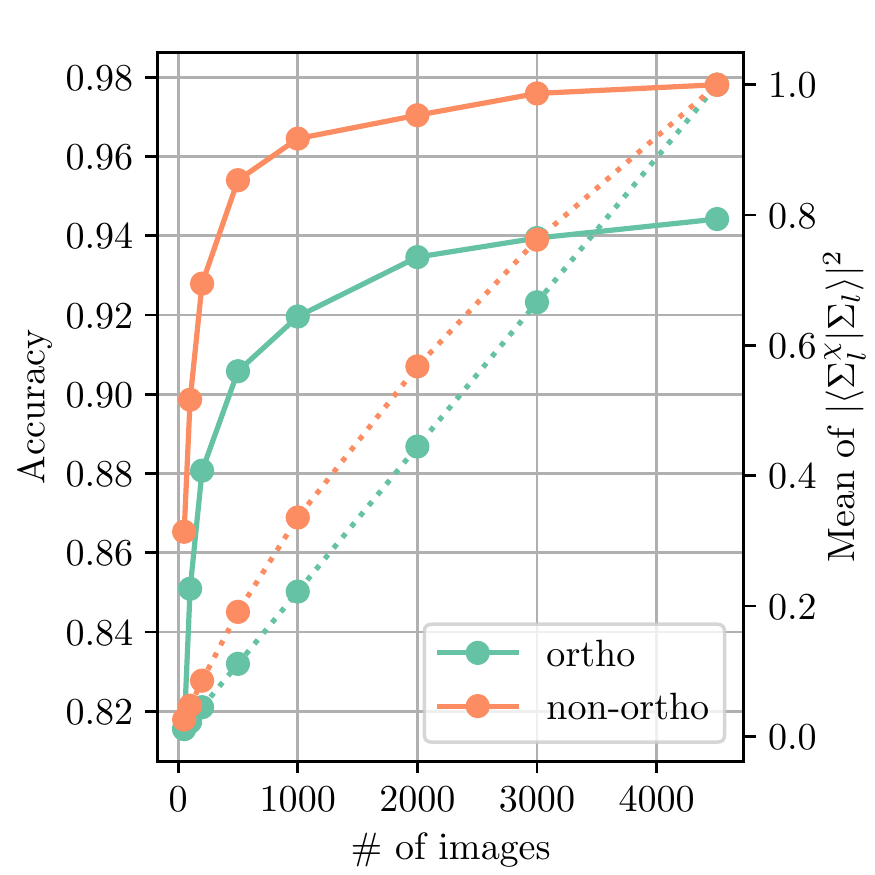}
\caption{The results of $\ket{\Sigma_l^\chi}$ formed by subset of the training data without performing compression.
The result for the orthonormal feature map $\phi(x) = [e^{i(3\pi/2)x}\cos(\frac{\pi}{2}x), e^{-i(3\pi/2)x}\sin(\frac{\pi}{2}x)]$ is colored in green.
The result for the non-orthonormal feature map $\phi(x) = [\cos(\frac{\pi x}{2}), \sin(\frac{\pi x}{2})]$  is colored in orange.
The solid lines indicate the test accuracy and the dashed lines indicate the mean of the square of overlaps to the exact state $\ket{\Sigma_l}$.
}
\label{fig: acc_overlap_nb}
\end{figure}

In Fig.~\ref{fig: acc_overlap_nb}, we show the result of taking partial training data to form $\ket{\Sigma_l^\chi}$ without performing compression.
The bond dimension $\chi$ is exactly the number of training data taken.
While in this case, we could still afford to exactly construct the states $\ket{\Sigma_l^\chi}$ of individual digit separately, in general the compression step is necessary for larger datasets and other tensor network states, e.g., PEPS.

\end{document}